\documentclass{article}
\pdfoutput=1 
\usepackage{spconf,amsmath,graphicx}
\usepackage{color}
\usepackage{times}
\usepackage{epsfig}
\usepackage{graphicx}
\usepackage{amsmath}
\usepackage{amssymb}
\usepackage{comment}
\usepackage[breaklinks=true,bookmarks=false]{}

\title{Turkish Presidential Elections TRT Publicity speech Facial Expression Analysis }

\name{H. Emrah Tasli , Paul Ivan}
\address{Vicarious Perception Technologies, Amsterdam, The Netherlands}

\begin{document}
%
\maketitle
\begin{abstract}

In this paper, facial expressions of the three Turkish presidential candidates Demirtas, Erdogan and Ihsanoglu (in alphabetical order) are analyzed during the publicity speeches featured at TRT (Turkish Radio and Television) on 03.08.2014. FaceReader is used for the analysis where 3D modeling of the face is achieved using the active appearance models (AAM). Over 500 landmark points are tracked and analyzed for obtaining the facial expressions during the whole speech. All source videos and the data are publicly available for research purposes.

\end{abstract}
\section{Introduction}
\label{sec:intro}

Remote monitoring of facial expressions is gaining popularity for many purposes. Psychological studies, market research and consumer analysis are some of the most common areas. In this study we intend to show the power of such methods during a public speech where three candidates of Turkish presidential elections (Demirtas, Erdogan, Ihsanoglu) are given equal amount of time to address the public from the national television on 03.08.2014. 

FaceReader is the world’s first tool capable of automatically analyzing facial expressions, providing users with an objective assessment of a person’s emotion \cite{Uyl2008}, \cite{Emrah2014ICIP}, \cite{VicarVision}. The RGB image of the person is used for the analysis in a machine learning framework where a training based facial landmark tracking is performed for obtaining a 3D model of the face. The 3D model is further used to obtain the facial expressions in a supervised learning framework.

In this study, the whole video of the public speech has been analyzed with the software to obtain a facial expression summary for the three candidates. The videos are analyzed fully automatic without using the speech. We do not intend to draw any kind of political stance or conclusions from the provided analysis.

\begin{figure}[]
\centering
\centerline{\includegraphics[width={\linewidth}]
{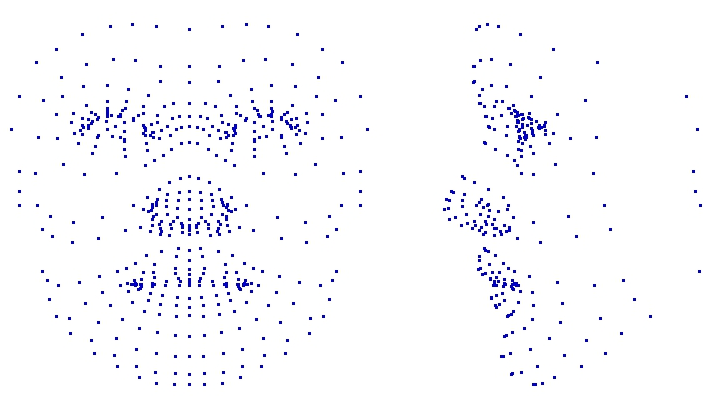}}
\caption{Frontal and lateral view of the obtained 3D facial landmark points on face}
\label{fig:ROI}
\vspace{-0.5cm}
\end{figure}

\section{Technique}

Active appearance models have been introduced by Cootes et al.\cite{cootes1998active} more than a decade ago for generating a parametric face model in relation to the annotated training dataset. The utilized facial analysis framework, commercially known as FaceReader \cite{Uyl2008}, uses an improved version of this methodology for obtaining very accurate appearance models for 3D facial landmark detection. The detected landmarks presented in Figure \ref{fig:ROI} are used to find the changes in the facial expressing of happy, sad, angry, surprised, scared, disgusted, contempt and neutral. In addition to facial expressions, valence (a measure of the attitude of the participant, positive vs negative), arousal (a measure of the activity of the participant), facial states (eyes/mouth opened/closed, eye brows lowered/neutral/raised), global gaze direction, action units, head pose and characteristics (gender, age, ethnicity and the presence of glasses, a beard and a mustache) are also measured during the experiment. More information could be found on the company website \cite{VicarVision}.

The videos are automatically analyzed during the publicity speech of the three candidates. The average of the facial expressions are presented in this study but further statistical analysis is possible in future studies. The parametric representation of the face with AAM, allows free head movement of the subject during the analysis. The facial landmark localization enables robust and accurate estimation of the facial expressions in the video.

\section{Experiments}
\label{sec:experiments}

\begin{figure}[]
\centerline{\includegraphics[width={\linewidth}]
{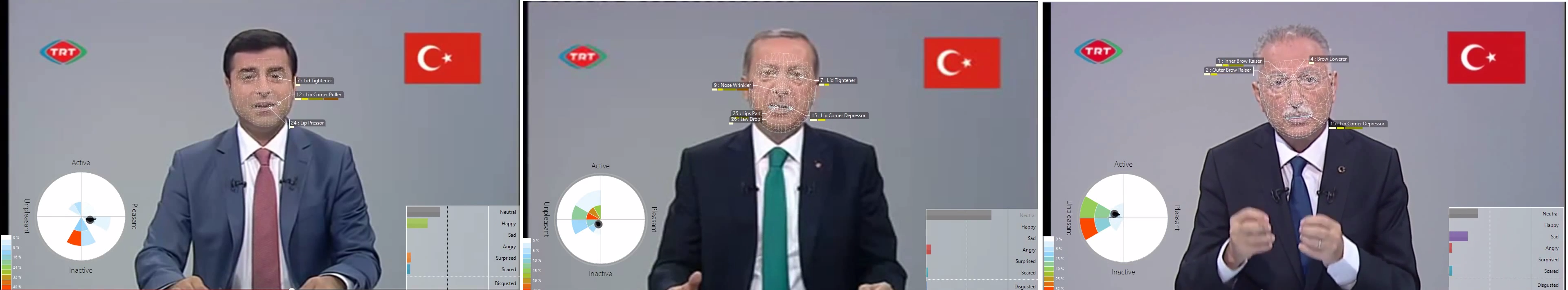}}
\caption{A frame from the analysis of the candidate videos, from left to right in alphabetical order: Demirtas, Erdogan, Ihsanoglu}
\label{candidates}
\end{figure}

\begin{figure}[]
\centerline{\includegraphics[width={\linewidth}]
{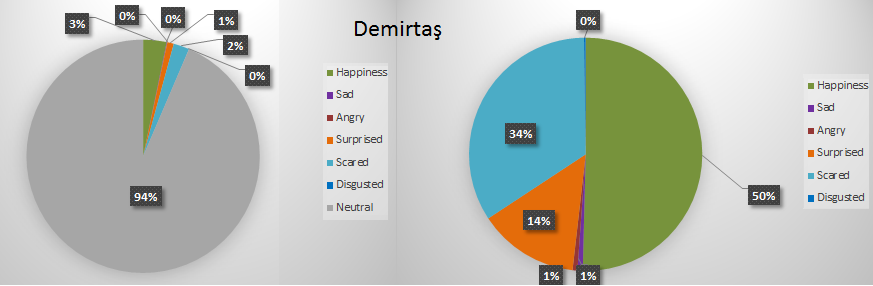}}
\vspace{0.2cm}
\centerline{\includegraphics[width={\linewidth}]
{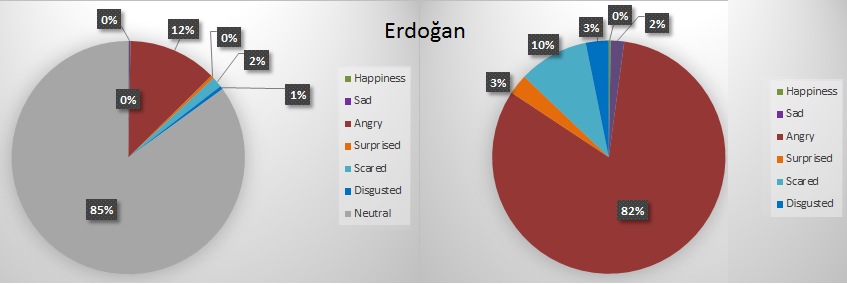}}
\vspace{0.2cm}
\centerline{\includegraphics[width={\linewidth}]
{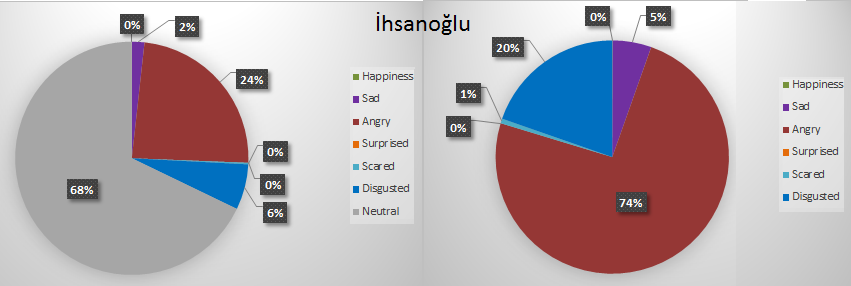}}
\caption{Left side shows the average facial expression of the candidates during the video. When the neutral state is discarded, results are shown on the right side.}
\label{elections}
\end{figure}

The three candidate public speech videos of length 10-15 min are analyzed. Figure \ref{candidates} shows a sample frame from the analysis screen. The resulting average facial expressions for the whole videos are presented in Figure \ref{elections}. Looking at the results, we would like to share some remarkable observations. Looking at the left column of Figure \ref{elections} one can conclude that the candidates do not use their facial expressions a lot and they generally have a neutral facial state during their speeches. As a comparison, Ihsanoglu stands outs as the candidate who uses his mimics the most with 32\%, Erdogan follows with 15\% and Demirtas is the one with the least usage of his mimics with 6\%.

When the neutral state (gray area on the left column) is removed from the analysis, we can see on the right column that there is a major difference between the candidates. It is observed that the common expression of Erdogan is anger with 82\% and fear is following that with 10\%. Similarly, the analysis of Ihsanoglu shows that, anger is the most common expression with 74\% and disgust follows that with 20\%. According to the analysis of Demirtas, the most common expression is happiness with 50\% and fear is following that with 34\%.

\section{Conclusion}
\label{sec:Conclusion}

This study shows the importance of an automatic facial expression analysis tool and the example scenario is selected as the public speeches of candidates of Turkish presidential elections that will take place on 10.08.2014. This is in less than a week at the time when paper is prepared. Such an analysis would definitely be more interesting during a real time discussion on a TV program where all the candidates are present. However, under the current conditions, this is the best possible unbiased way to have a comparative analysis during the publicity speech where TRT is officially liable to broadcast. We hope to make a future study during a real time broadcast with all the candidates present.

The website of the software and the video links are presented. The data is publicly available for research purposes. The analysis can be validated using the videos on the project website \cite{FaceReaderAnalysisBlog} and FaceReader 6.0 software. All the results are obtained scientifically and they do not represent or reflect any personal view or political opinion. The purpose of this study is to analyze the facial expressions of the presidential candidates during their publicity speeches.


\bibliographystyle{IEEEbib}
\bibliography{../emrah}

\end{document}